\documentclass[conference]{IEEEtran}
\IEEEoverridecommandlockouts

\usepackage{cite}
\usepackage{amsmath,amssymb,amsfonts}
\usepackage{algorithmic}
\usepackage{graphicx}
\usepackage{textcomp}
\usepackage{longtable}
\usepackage{multirow}
\usepackage{xcolor}
\usepackage{url}
\usepackage[russian,british]{babel}
\def\BibTeX{{\rm B\kern-.05em{\sc i\kern-.025em b}\kern-.08em
    T\kern-.1667em\lower.7ex\hbox{E}\kern-.125emX}}
\begin{document}

\title{Evaluating LLM Prompts for Data Augmentation in Multi-label Classification of Ecological Texts

\thanks{This study was supported by the Ministry of Science and Higher Education of the Russian Federation within the framework of the Carbon Measurement Test Area in Tyumen' Region (FEWZ-2024-0016).}
}

\author{\IEEEauthorblockN{Anna Glazkova}
\IEEEauthorblockA{\textit{Carbon Measurement Test Area in Tyumen' Region } \\
\textit{(FEWZ-2024-0016),}\\
\textit{University of Tyumen}\\
Tyumen, Russia \\
0000-0001-8409-6457}
\and
\IEEEauthorblockN{Olga Zakharova}
\IEEEauthorblockA{\textit{Carbon Measurement Test Area in Tyumen' Region } \\
\textit{(FEWZ-2024-0016),}\\
\textit{University of Tyumen}\\
Tyumen, Russia \\
0000-0002-1404-4915}
}

\maketitle

\begin{abstract}
Large language models (LLMs) play a crucial role in natural language processing (NLP) tasks, improving the understanding, generation, and manipulation of human language across domains such as translating, summarizing, and classifying text. Previous studies have demonstrated that instruction-based LLMs can be effectively utilized for data augmentation to generate diverse and realistic text samples. This study applied prompt-based data augmentation to detect mentions of green practices in Russian social media. Detecting green practices in social media aids in understanding their prevalence and helps formulate recommendations for scaling eco-friendly actions to mitigate environmental issues. We evaluated several prompts for augmenting texts in a multi-label classification task, either by rewriting existing datasets using LLMs, generating new data, or combining both approaches. Our results revealed that all strategies improved classification performance compared to the models fine-tuned only on the original dataset, outperforming baselines in most cases. The best results were obtained with the prompt that paraphrased the original text while clearly indicating the relevant categories.
\end{abstract}


\section{Introduction}

Text classification is an important task in natural language processing, with many practical applications. In real-world classification tasks, training data for classification models is often imbalanced. This negatively affects classification performance, as the model tends to struggle with identifying minority classes \cite{johnson2019survey}. One way to address class imbalance is to increase the number of texts in minority classes through data augmentation. Data augmentation is the process of constructing synthetic data from an available dataset \cite{shorten2021text}. Due to the impressive advancements in large language models~(LLMs), recent studies on data augmentation have increasingly utilized these models~\cite{sufi2024generative,chen2023empirical}. Researchers employed various approaches to prompt-based data augmentation with LLMs, including paraphrasing the original text, generating new samples, and combining these approaches.

This paper explored and compared several prompting strategies for data augmentation to improve the performance of multi-label text classifiers. Since it is unclear which of the approaches to prompt-based data augmentation is more beneficial, this study compares existing strategies. The experiment focused on an imbalanced dataset to detect mentions of green practices in Russian-language social media texts. Green practices are everyday actions that harmonize the relationship between humans and the environment \cite{zakharova2021green}. The importance of this task stems from the need to broaden current understanding of green practices in society and to track the spread of these practices \cite{glazkova2022detecting}. To effectively implement green practices, it's important to know what practices already exist, how common they are, and who initiates and supports them. However, awareness of their prevalence is limited due to the time-consuming nature of traditional data collection methods such as surveys and interviews \cite{zakharova2021greenA}. At the same time, social media contains a large amount of unstructured text data on environmental issues. Automatically analyzing this data can help faster gather and process information, providing insights into the spread of various green practices.

Since green practices are unevenly spread, mentions of some types of practices are rare. Due to the small number of text examples mentioning these practices, machine learning models struggle to detect the rarer ones. One of the solutions to the problem of imbalanced datasets can be data augmentation.

The contributions of this study can be summarized as follows. Firstly, we experimented with four prompt-based data augmentation strategies to improve multi-label classification performance, namely rewriting existing texts, generating new samples, and two strategies combining text rewriting and new data generation. We showed that paraphrasing the original text considering the relevant categories led to performance increase in our experiments. Secondly, we explored these data augmentation strategies for the first time on a Russian dataset using a model specifically trained for the Russian language.

The rest of the paper is organized as follows. Section 2 contains a brief review of related work. Section 3 describes the dataset and methods used. Section 4 provides and discusses the experimental results. Section 5 concludes this paper.

\section{Related Work}

Data augmentation aims to construct synthetic data from an available dataset to reduce the class imbalance and increase the classification performance. Previously, augmentation methods included text transformation techniques such as token deletion, insertion, and replacement \cite{wei2019eda,karimi2021aeda}, and back-translation~\cite{sennrich1511improving}. Recent methods utilized pre-trained language models, which enable the generation of more natural texts due to the linguistic knowledge embedded within them \cite{kumar2020data,wu2019conditional}. Some studies have employed instruct-based LLMs for data augmentation~(see Table \ref{summary}). For example, Dai et al. \cite{dai2023auggpt} proposed a data augmentation approach based on ChatGPT that rephrases each sentence in the training samples into multiple conceptually similar but semantically different samples. Piedboeuf and Langlais \cite{piedboeuf2023chatgpt} used ChatGPT for data augmentation both with paraphrasing and with zero-shot generation, and compared it to seven other algorithms. The results obtained varied depending on the dataset used, but both approaches involving ChatGPT demonstrated high performance compared to the baselines. Zhao et al. \cite{zhao2024improving} proposed an approach to combining paraphrasing and generating strategies through rewriting of the generated text.

A brief review of existing work shows that most studies on the use of prompt-based augmentation for improving text classification rely on models from the Open AI family. Researchers employ various strategies based on paraphrasing the original text, generating new text, or combining these strategies. To the best of the authors' knowledge, such experiments have not yet been conducted for Russian texts and multi-label datasets. Additionally, the effectiveness of open-source models for text augmentation remains underexplored. This study aims to address this research gap.

\begin{table*}[h!]
\centering
\caption{Summary of recent studies that used prompt-based data augmentation to improve text classification performance.\label{summary}}
\begin{tabular}{|p{0.7cm}|p{1.5cm}|p{4cm}|p{2cm}|p{2cm}|p{5.5cm}|}
\hline
Paper & Classification task & Domain & Model & Language & Prompting strategy \\ \hline
\cite{yoo2021gpt3mix} & Multiclass & Multiple (7 datasets) & GPT-3 & English & Generating new samples using given categories and the examples from the original dataset \\ \hline
\cite{dai2023auggpt} & Multiclass & Multiple (3 datasets) & ChatGPT & English & Paraphrasing \\ \hline
\cite{sarker2023medical} & Multiclass & Medical & ChatGPT & English & Paraphrasing \\ \hline
\cite{piedboeuf2023chatgpt} & Multiclass, binary& Multiple (5 datasets) & ChatGPT & English & Paraphrasing, generating new samples \\ \hline
\cite{cohen2023enhancing} & Binary& Social media (hate speech detection) & GPT-3 & English & Paraphrasing \\ \hline
\cite{shushkevich2023improving} & Multiclass  & News (fake news detection) & ChatGPT & English, German & Paraphrasing \\ \hline
\cite{wozniak2023big} & Multiclass  & Reviews, crowd-sourced annotations & GPT-3.5 & English & Paraphrasing, generating new samples using given categories and the examples from the original dataset \\ \hline
\cite{moller2024parrot} & Multiclass, binary & Multiple (10 datasets) & ChatGPT, Llama & English, Danish & Generating new samples using given categories and the examples from the original dataset \\ \hline
\cite{zhao2024improving} & Multiclass & News, ecological & ChatGPT & English & Paraphrasing, generating, combining paraphrasing and generating through rewriting of the generated sample\\ \hline
\end{tabular}
\end{table*}

\section{Methods}

\subsection{Data}

\begin{figure}[]
\center{\includegraphics[width=1\linewidth]{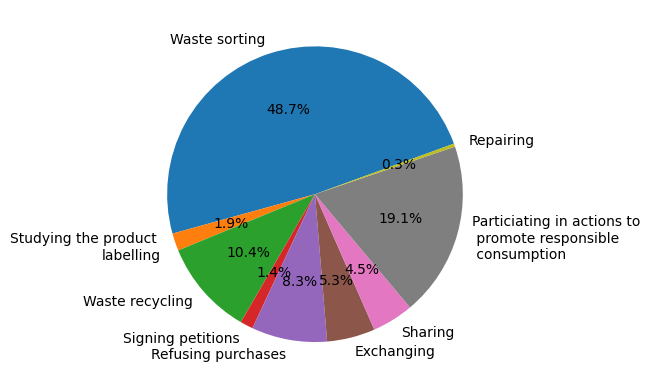}} \\
\caption{The distribution of practices in GreenRu.}
\label{ris:practices}
\end{figure}

\begin{figure}[]
\center{\includegraphics[width=1\linewidth]{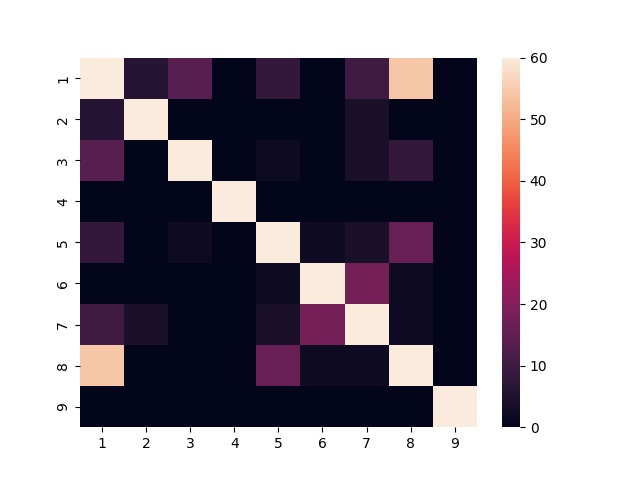}} \\
\caption{The mutual occurrence of practices in GreenRu. 1 - waste sorting, 2 - studying the product labeling, 3 - waste recycling, 4 - signing petitions, 5 - refusing purchases, 6 - exchanging, 7 - sharing, 8 - participating in actions to promote responsible consumption, 9 - repairing.}
\label{ris:occur}
\end{figure}

This study used GreenRu\footnote{\url{https://github.com/green-solutions-lab/GreenRu}}, a Russian social media dataset with labeled mentions of green practices \cite{zakharova2024greenru}. GreenRu has a sentence-level multi-label markup and consists of 1326 posts collected in Russian online communities. We used sentences as input texts for classifiers. The dataset was annotated using nine types of green waste practices~\cite{zakharova2022importance}: 1) \textbf{waste sorting}, which involves separating waste by type; 2) \textbf{studying the product labeling} to identify packaging as a type of waste; 3) \textbf{waste recycling}, which refers to converting waste materials into reusable resources for future production; 4) \textbf{signing petitions} to influence authorities; 5) \textbf{refusing purchases}, where individuals choose not to buy products or services that negatively impact the environment, thus reducing consumption and their environmental footprint; 6) \textbf{exchanging}, which entails giving away an unnecessary item or service in return for something desired; 7) \textbf{sharing}, where multiple individuals use a single item for a fee or free of charge; 8) \textbf{participating in actions to promote responsible consumption}, such as workshops, festivals, and lessons focused on encouraging reduced consumption; 9) \textbf{repairing}, which involves restoring the functionality of items instead of discarding them. The distribution of green practices across the entire dataset and the mutual occurrence of green practices are illustrated in Figures~\ref{ris:practices} and \ref{ris:occur}. The detailed dataset statistics is presented in Table~\ref{stat}. The symbol $\pm$ indicates standard deviation values.

\begin{table}[]
\centering
\caption{Dataset statistics.\label{stat}}
\begin{tabular}{|p{4cm}|p{1.8cm}|p{1.8cm}|}\hline
Characteristic & Training subset & Test subset \\\hline
Number of posts & 913 & 413 \\
Number of sentences with multi-label markup & 2442 & 1058 \\
Avg symbols per post & 880.05$\pm$751.46 & 908.53$\pm$761.06 \\
Avg symbols per sentence & 111.35$\pm$101.23 & 114.99$\pm$101.73 \\\hline
\multicolumn{3}{|c|}{Number of mentions per practice} \\\hline
Waste sorting& 1275 &560  \\
Studying the product labeling&  55& 17 \\
Waste recycling & 272 &121  \\
Signing petitions &  22&31  \\
Refusing purchases &  236&75  \\
Exchanging &  146&52  \\
Sharing &  109&62  \\
Participating in actions to promote responsible consumption &  510&209  \\
Repairing &  10&3  \\\hline
\end{tabular}
\end{table}

\subsection{Prompting Strategies}

T-lite-instruct-0.1 (\textbf{T-lite})\footnote{\url{https://huggingface.co/AnatoliiPotapov/T-lite-instruct-0.1}} was used as an instruction-based LLM to generate text samples for data augmentation. T-lite is a model designed specifically for the Russian language, enabling the creation of large language model applications in Russian. The proportion of data in Russian for pre-training T-lite was 85\%. We used the maximum number of tokens to generate equal to 400 and a temperature of 0.5.

Four prompting strategies were used to obtain augmented data from T-lite:
\begin{itemize}
    \item Paraphrasing text, $P_{text}$: 
    \begin{itemize}
        \item \selectlanguage{russian}\texttt{Перефразируй текст: [TEXT]}\selectlanguage{british} (\textit{Paraphrase the text: [TEXT]}). 
    \end{itemize}
    
    \item Generating new sample based on given categories, $G_{topics}$: 

    \begin{itemize}
        \item \selectlanguage{russian}\texttt{Напиши короткий пост для экологического сообщества в социальной сети, относящийся к тематикам: [TOPICS]}\selectlanguage{british} (\textit{Write a short post for an environmental community on social media, related to the following topics: [TOPICS]}).
    \end{itemize}

    \item Paraphrasing text with a focus on its relation to the given categories, $P_{text+topics}$: 
    
    \begin{itemize}
        \item \selectlanguage{russian}\texttt{Перефразируй текст с учетом того, что он относится к следующим тематикам: [TOPICS]. Исходный текст: [TEXT]}\selectlanguage{british} (\textit{Paraphrase the text considering its relevance to the following topics: [TOPICS]. Original text: [TEXT]}).
    \end{itemize}
    
    \item Generating new sample based on given categories and the textual example, $G_{topics+text}$: 
    
    \begin{itemize}
        \item \selectlanguage{russian}\texttt{Напиши короткий пост для экологического сообщества в социальной сети, относящийся к тематикам: [TOPICS]. Например: [TEXT]}\selectlanguage{british} (\textit{Write a short post for an environmental community on social media, related to the following topics: [TOPICS]. For example: [TEXT]}).
    \end{itemize}

\end{itemize}

[TEXT] and [TOPICS] are an original text from the training set and its labels in a textual form respectively. The example of prompts and generated samples are shown in Table~\ref{table_example}, the translation into English is italicized. All generated texts are available publicly\footnote{\url{https://github.com/green-solutions-lab/GreenRu/tree/main/ai-generated}}.

\begin{table*}[]
\centering
\scriptsize
\caption{Example of generated samples.\label{table_example}}
\addtolength{\tabcolsep}{-1pt}
\begin{tabular}{|p{1cm}p{16cm}|}
\hline
\multicolumn{1}{|p{1cm}|}{Text} &  \selectlanguage{russian}В целях создания сказочного настроения и поддержки всех начинающих волшебников и волшебниц мы объявляем конкурс творческих работ "В новый год без старья: красота из вторсырья".\selectlanguage{english} \textit{To create a magical atmosphere and to support all aspiring wizards and witches, we are announcing a creative contest entitled "Into the New Year without the Old: Beauty from Recyclables".}\\ \hline
\multicolumn{1}{|p{1cm}|}{Topics} &  waste recycling, participating in actions to promote responsible consumption\\ \hline\hline
\multicolumn{2}{|c|}{$P_{text}$} \\ \hline
\multicolumn{1}{|p{1cm}|}{Prompt} & \selectlanguage{russian}Перефразируй текст: В целях создания сказочного настроения и поддержки всех начинающих волшебников и волшебниц мы объявляем конкурс творческих работ "В новый год без старья: красота из вторсырья".\selectlanguage{english} \textit{Paraphrase the text: To create a magical atmosphere and to support all aspiring wizards and  witches, we are announcing a creative contest titled "Into the New Year without the Old: Beauty from Recyclables".}
\\ \hline
\multicolumn{1}{|p{1cm}|}{Result} &  \selectlanguage{russian}Для того чтобы вдохновить и создать атмосферу волшебства, а также помочь юным магам и чародейкам, мы организуем конкурс "Новый год без отходов: чудеса из переработанных материалов". Ваша задача — проявить фантазию и сделать что-то необычное из старых вещей, превратив их в настоящие произведения искусства!\selectlanguage{english}

\textit{To inspire and to create a magical atmosphere, as well as support young wizards and sorceresses, we are organizing a contest entitled "A Waste-Free New Year: Wonders from Recycled Materials." Your task is to unleash your creativity and turn old items into true works of art!}\\ \hline
\multicolumn{2}{|c|}{$G_{topics}$} \\ \hline
\multicolumn{1}{|p{1cm}|}{Prompt} & \selectlanguage{russian}Напиши короткий пост для экологического сообщества в социальной сети, относящийся к тематикам: переработка отходов, участие в акциях по продвижению ответственного потребления.\selectlanguage{english} \textit{Write a short post for an environmental community on social media, related to the following topics: waste recycling, participating in actions to promote responsible consumption.}\\ \hline
\multicolumn{1}{|p{1cm}|}{Result} &  \selectlanguage{russian}Ваше участие в защите планеты — это важный шаг!

Сегодня мы хотим поделиться с вами несколькими простыми, но действенными способами, как каждый из нас может внести вклад в сохранение природы.

1. Переработка отходов:
   - Пластиковые бутылки и пакеты: Вместо того чтобы выбрасывать их, сдавайте на переработку. В некоторых городах есть пункты приема, а также можно использовать многоразовые контейнеры.
   - Бумага и картон: Не выбрасывайте макулатуру, а складывайте её в специальные контейнеры или используйте для упаковки.
   - Электроника: Старые телефоны, ноутбуки и батарейки можно сдать в специальные пункты утилизации.

2. Активное участие в акциях:
   - Участвуйте в субботниках: Вместе с друзьями и соседями вы можете организовать уборку местных парков\selectlanguage{english}

\textit{Your participation in protecting the planet is an important step!}

\textit{Today, we want to share a few simple yet effective ways each of us can contribute to preserving nature.}

\textit{1. Recycling waste:}
   \textit{- Plastic bottles and bags: Instead of throwing them away, recycle them. Some cities have collection points, and you can also use reusable containers.}
   \textit{- Paper and cardboard: Don’t throw away paper waste—collect it in special bins or use it for packaging.}
   \textit{- Electronics: Old phones, laptops, and batteries can be taken to special recycling points.}

\textit{2. Active participation in campaigns:}
   \textit{- Join community clean-ups: You and your friends or neighbors can organize clean-ups of local parks.}
\\ \hline

\multicolumn{2}{|c|}{$P_{text+topics}$} \\ \hline
\multicolumn{1}{|p{1cm}|}{Prompt} &  \selectlanguage{russian}Перефразируй текст с учетом того, что он относится к следующим тематикам: переработка отходов, участие в акциях для продвижения ответственного потребления. Исходный текст: В целях создания сказочного настроения и поддержки всех начинающих волшебников и волшебниц мы объявляем конкурс творческих работ "В новый год без старья: красота из вторсырья".\selectlanguage{english} \textit{Paraphrase the text considering its relevance to the following topics: waste recycling, participating in actions to promote responsible consumption. Original text: To create a magical atmosphere and to support all aspiring wizards and  witches, we are announcing a creative contest entitled "Into the New Year without the Old: Beauty from Recyclables".}
\\ \hline
\multicolumn{1}{|p{1cm}|}{Result} &  \selectlanguage{russian}В преддверии волшебного праздника, чтобы вдохновить и поддержать всех, кто верит в чудеса и творчество, мы запускаем конкурс "Новогодний фейерверк из вторичных ресурсов". Ваша задача — создать необычные и яркие поделки, используя материалы, которые уже служили людям, и тем самым внести вклад в экологию и продемонстрировать, что красота может быть рождена из старых вещей. Пусть ваша фантазия и мастерство помогут превратить отходы в настоящие произведения искусства!\selectlanguage{english}

\textit{As the magical holiday approaches, to inspire and to support everyone who believes in creativity and miracles, we are launching the contest "New Year's Fireworks from Recycled Materials." Your task is to create unique and vibrant crafts using materials that have already served people, thus contributing to ecology and demonstrating that beauty can be born from old things. Let your imagination and skills transform waste into true works of art!}\\ \hline
\multicolumn{2}{|c|}{$G_{topics+text}$} \\ \hline
\multicolumn{1}{|p{1cm}|}{Prompt} &  \selectlanguage{russian}Напиши короткий пост для экологического сообщества в социальной сети, относящийся к тематикам: переработка отходов, участие в акциях по продвижению ответственного потребления. Например: В целях создания сказочного настроения и поддержки всех начинающих волшебников и волшебниц мы объявляем конкурс творческих работ "В новый год без старья: красота из вторсырья".\selectlanguage{english} \textit{Write a short post for an environmental community on social media, related to the following topics: waste recycling, participating in actions to promote responsible consumption. For example: To create a magical atmosphere and to support all aspiring wizards and  witches, we are announcing a creative contest entitled "Into the New Year without the Old: Beauty from Recyclables".}\\ \hline
\multicolumn{1}{|p{1cm}|}{Result} & \selectlanguage{russian}Встречайте Новый год с чистой совестью и яркими идеями!

Дорогие друзья и волшебники экологии! Приближается волшебный праздник, и мы хотим, чтобы он стал не только радостным, но и экологически ответственным. В этом году мы решили объединить два мира: магию и заботу о природе, и приглашаем вас принять участие в нашем конкурсе "В новый год без старья: красота из вторсырья".

Что нужно сделать:

1. Создайте что-то волшебное: Поделки, украшения, декорации, которые будут сделаны из переработанных материалов. Это могут быть старые журналы, картонные коробки, пластиковые бутылки, стеклянные банки и многое другое.
2. Фантазия без границ: Вдохновляйтесь сказками, мифами и собственными идеями. Важно, чтобы ваше творчество было не только красивым, но и полезным для планеты.
3. Поделитесь историей: Расскажите, как вы нашли вдохновение, какие материалы использовали и почему это важно для вас. Ваши истории помогут другим понять, что каждый может внести вклад в защиту природы.

Сроки и условия:
- Прием работ: с 1 декабря по 15 января.
- Голосование: с 16 по 31 января.
- Награждение: 1 февраля\selectlanguage{english}

\textit{Welcome the New Year with a clear conscience and bright ideas! 
Dear friends and eco-wizards! The magical holiday is approaching, and we want it to be not only joyful but also environmentally responsible. This year, we decided to combine two worlds: magic and care for nature, and we invite you to participate in our contest, "Into the New Year Without the Old: Beauty from Recyclables."}

\textit{What you need to do:}
\textit{1. Create something magical: Crafts, decorations, or ornaments made from recycled materials. You can use old magazines, cardboard boxes, plastic bottles, glass jars, and much more.}
\textit{2. Unlimited creativity: Be inspired by fairy tales, myths, and your own ideas. The key is for your creation to be not only beautiful but also beneficial to the planet.}
\textit{3. Share your story: Tell us how you found your inspiration, what materials you used, and why this is important to you. Your stories will help others understand that everyone can contribute to protecting the environment.}

\textit{Timeline and rules:}
\textit{- Submission period: December 1 to January 15.}
\textit{- Voting: January 16 to January 31.}
\textit{- Awards ceremony: February 1.}
\\ \hline
\end{tabular}
\end{table*}

\subsection{Sampling Strategies}

\begin{figure}[h]
\begin{minipage}[h]{1\linewidth}
\center{\includegraphics[width=1\linewidth]{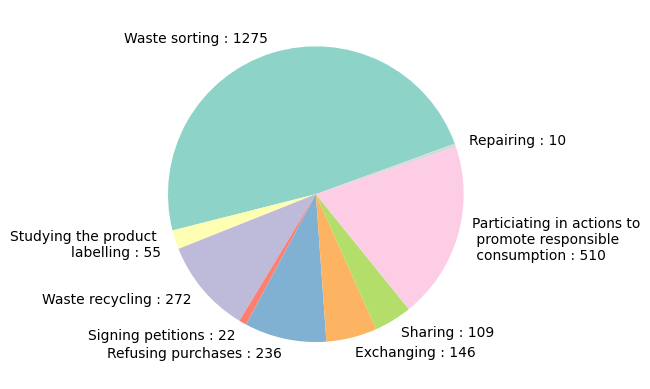} \\ a) Original data}
\end{minipage}
\vfill
\begin{minipage}[h]{1\linewidth}
\center{\includegraphics[width=1\linewidth]{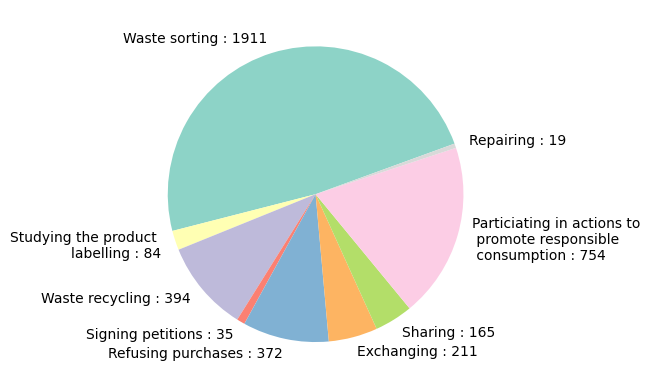} \\ b) $\times$ 1.5}
\end{minipage}
\vfill
\begin{minipage}[h]{1\linewidth}
\center{\includegraphics[width=1\linewidth]{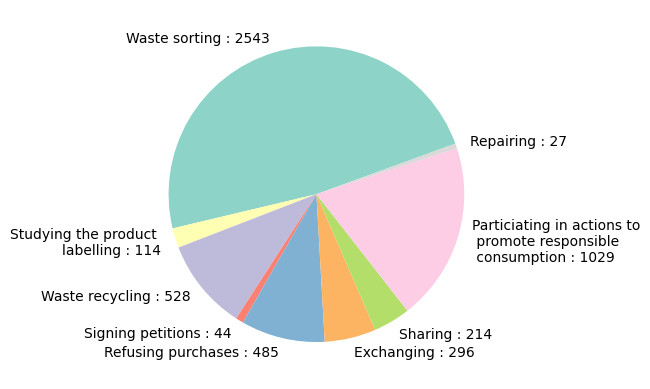} \\ c) $\times$ 2}
\end{minipage}
\caption{The number of mentions per practice in the training set.}
\label{ris:aug}
\end{figure}

This study used two sampling strategies that differed only in the amount of added data. The augmentation process included the following steps. Firstly, we randomly selected a sentence from the original training set, then applied an augmentation technique to it, and, finally, added the new sentence to the training set. Sampling of new data continued until the size of the training set increased by 1.5 times (3663 sentences) and 2 times (4884 sentences) for the first and second strategies, respectively. The number of mentions per practice in the original and augmented training sets is presented in Figure~\ref{ris:aug}. Since we randomly selected a sentence from the original training set, the ratio of practices after augmentation resembles the original ratio. However, the number of examples for each practice, including rare ones, has increased.

\subsection{Baselines}

The considering prompting strategies were compared with two data augmentation baselines:

\begin{itemize}
    \item \textbf{Random duplication}. For this technique, we did not generate new samples for original sentences and simply duplicated random sentences from the training set without any processing.
    \item \textbf{Back translation} \cite{sennrich1511improving}. This technique uses back translating phrases between any two languages. We utilized the BackTranslation library\footnote{\url{https://pypi.org/project/BackTranslation}} based on Google Translate and English as a target language.
\end{itemize}

\subsection{Classifiers}

Two fine-tuned language models were used as classifiers to assess the performance of data augmentation techniques.

\begin{itemize}
    \item \textbf{ruELECTRA}\footnote{\url{https://huggingface.co/ai-forever/ruElectra-large}} \cite{zmitrovich2024family}, the model that uses the ELECTRA architecture \cite{clark2020electra}. ruELECTRA was pre-trained on a large collection of Russian texts from various publicly available resources, which represented diverse domains. We used the ruELECTRA-large version of the model.
    \item \textbf{ruBERT}\footnote{\url{https://huggingface.co/DeepPavlov/rubert-base-cased}} \cite{kuratov2019adaptation}, the first adaptation of the BERT architecture \cite{devlin-etal-2019-bert} for Russian pre-trained on the Russian part of Wikipedia and news data.
\end{itemize}

Both models were fine-tuned for five epochs using a maximum sequence length of 256 tokens, the AdamW optimizer, a learning rate equal to 4e-5, and a batch size equal to eight. The models were fine-tuned in a multi-label manner, i. e. the target for a single example from the dataset is a list of $n$ distinct binary labels. Each classifier used a transformer model with a classification layer that had $n$ output neurons, one for each label. The models were implemented using the Simple Transformers library \cite{rajapakse2019simpletransformers}.

\subsection{Evaluation Metric}

Due to the multi-label annotation of GreenRu, we used the multi-label F1-score as an evaluation metric. The multi-label F1-score calculates the scores of each class and takes the average of them.

\section{Results and Discussion}

\begin{figure}[]
\center{\includegraphics[width=1\linewidth]{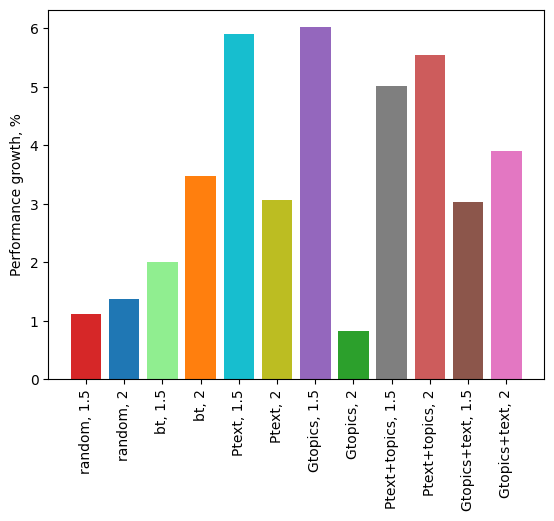}} \\
\caption{Performance growth, \% (ruELECTRA).}
\label{ris:electra}
\end{figure}

\begin{figure}[]
\center{\includegraphics[width=1\linewidth]{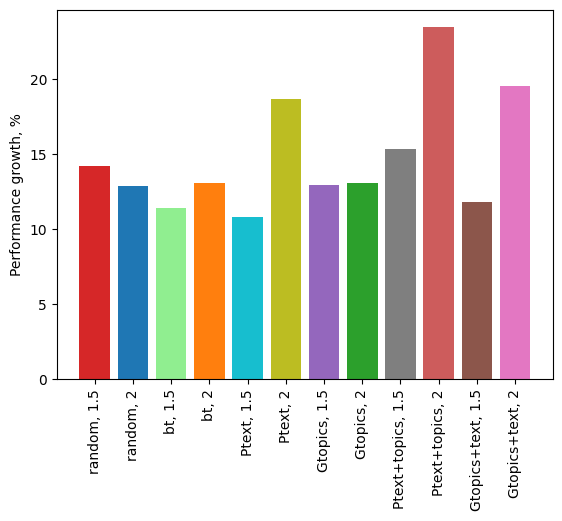}} \\
\caption{Performance growth, \% (ruBERT).}
\label{ris:bert}
\end{figure}

Table \ref{table_results} presents the evaluation results. The results are averaged across three model runs, the corresponding standard deviation values are marked with the symbol $\pm$. The best results for each sampling strategy (an increase in the training set by 1.5 or 2 times) are shown in bold. The second highest results for each sampling strategy are highlighted in italics and bold. The performance growth relative to models fine-tuned on the original data is shown in Figures \ref{ris:electra} and \ref{ris:bert}.

The results reveal that data augmentation increased the multi-label F1-score across all augmentation techniques. Overall, the results of data augmentation using instruction-based LLMs were higher than the baselines. For ruELECTRA, the best results were demonstrated using the $G_{topics}$ and $P_{text+topics}$ prompts. The $G_{topics+text}$ and $P_{text}$ showed second highest results. For ruBERT, both best scores were obtained using the $P_{text+topics}$ prompt. Second highest results were achieved with the $G_{topics+text}$ prompt and random duplication. Thus, in three out of four cases, the $P_{text+topics}$ prompt demonstrated the highest performance in terms of the multi-label F1-score. In our experiments, using a prompt that involves paraphrasing the text with a clear indication of the categories it belongs to proved effective for the task of multi-label text classification. At the same time, we did not observe a significant difference in the results of the other three prompts. All of them contributed to improving classification performance compared to the models fine-tuned on the original dataset, and in most cases, they outperformed the baselines.

\begin{table}[]
\centering
\caption{Results, \textit{\%}.\label{table_results}}
\begin{tabular}{|p{6cm}|p{1.5cm}|}
\hline
Model & F1-score \\ \hline\hline
ruELECTRA, original data &  69.96$\pm$5.76\\ \hline
\qquad+ random duplication ($\times$1.5)&  70.74$\pm$4.33\\ \hline
\qquad+ random duplication ($\times$2) &  70.92$\pm$1.49\\ \hline
\qquad+ back translation ($\times$1.5) &  71.36$\pm$4.95\\ \hline
\qquad+ back translation ($\times$2) &  72.39$\pm$0.75\\ \hline
\qquad+ $P_{text}$ ($\times$1.5)& \textbf{\textit{74.09$\pm$0.86}} \\ \hline
\qquad+ $P_{text}$ ($\times$2) &  72.11$\pm$2.44\\ \hline
\qquad+ $G_{topics}$ ($\times$1.5)& \textbf{74.17$\pm$0.47} \\ \hline
\qquad+ $G_{topics}$ ($\times$2) &  70.54$\pm$0.95\\ \hline
\qquad+ $P_{text+topics}$ ($\times$1.5)& 73.47$\pm$2.07 \\ \hline
\qquad+ $P_{text+topics}$ ($\times$2) &  \textbf{73.84$\pm$0.86}\\ \hline
\qquad+ $G_{topics+text}$ ($\times$1.5)&  72.08$\pm$2.30\\ \hline
\qquad+ $G_{topics+text}$ ($\times$2) &  \textbf{\textit{72.69$\pm$0.68}}\\ \hline\hline
ruBERT, original data &  58.16$\pm$2.52\\ \hline
\qquad+ random duplication ($\times$1.5)&  \textbf{\textit{66.41$\pm$1.16}}\\ \hline
\qquad+ random duplication ($\times$2) &  65.63$\pm$1.10\\ \hline
\qquad+ back translation ($\times$1.5) &  64.80$\pm$0.44\\ \hline
\qquad+ back translation ($\times$2) &  65.76$\pm$0.76\\ \hline
\qquad+ $P_{text}$ ($\times$1.5)&  64.42$\pm$0.53\\ \hline
\qquad+ $P_{text}$ ($\times$2) & 69.02$\pm$3.92 \\ \hline
\qquad+ $G_{topics}$ ($\times$1.5)&  65.68$\pm$0.96\\ \hline
\qquad+ $G_{topics}$ ($\times$2) & 65.75$\pm$1.58 \\ \hline
\qquad+ $P_{text+topics}$ ($\times$1.5)& \textbf{67.07$\pm$2.45} \\ \hline
\qquad+ $P_{text+topics}$ ($\times$2) & \textbf{71.78$\pm$1.61} \\ \hline
\qquad+ $G_{topics+text}$ ($\times$1.5)&  65.00$\pm$0.56\\ \hline
\qquad+ $G_{topics+text}$ ($\times$2) & \textbf{\textit{69.50$\pm$3.01}} \\ \hline
\end{tabular}
\end{table}

Table \ref{bertscore} shows the average values of ROUGE-1 \cite{lin-2004-rouge}, ROUGE-L, and BERTScore \cite{zhangbertscore} for the 50 random generated samples and original data. These metrics show how much the generated sample corresponds to the original text and similarly how much the augmented text retains the general meaning of the original text. As expected, the highest similarity to the original text is shown by the $P_{text}$ and $P_{text+topics}$ prompts, which are based on paraphrasing. However, the scores for $P_{text}$ are significantly higher than for $P_{text+topics}$. Strategies based on the generation of new samples show less similarity to the original text. The lowest similarity is demonstrated with $G_{topics}$.

\begin{table}[]
\centering
\caption{The values of ROUGE-1, ROUGE-L, and BERTScore for the generated samples and original data, \%.\label{bertscore}}
\begin{tabular}{|c|c|c|c|}\hline
Prompt & ROUGE-1 & ROUGE-L & BERTScore \\\hline
$P_{text}$ & 32.58 & 30.50 & 76.88 \\\hline
$G_{topics}$ & 4.03 & 3.77 & 56.40 \\\hline
$P_{text+topics}$ & 18.94 & 17.65 & 70.40 \\\hline
$G_{topics+text}$ & 7.60 & 7.10 & 60.66\\\hline
\end{tabular}
\end{table}

\section{Conclusion}

In this study, we compared four strategies for prompt-based data augmentation using the task of multi-label text classification in ecological domain. These strategies encompass various approaches to combining the original text with information about the categories it belongs to generating new samples. We experimented with two transformer-based classifiers using different amounts of newly generated samples. The experimental results indicated that employing a prompt that clearly identifies the associated categories of the text was effective for text classification. This study suggests guidance on the best strategies for using LLMs to improve text classification models. It could be especially beneficial for multi-label text classification tasks that experience significant class imbalance challenges.

This study was limited to the random selection of texts from the training subset for augmentation. Further research will focus on comparing sampling strategies to generate the optimal amount of texts related to rare categories.

\section*{Acknowledgment}

We are grateful to Nadezhda Zhuravleva (Center for Academic Writing ``Impulse'', University of Tyumen) for her assistance with the English language.

\bibliographystyle{IEEEtran}
\bibliography{IEEE-conference-template-062824}

\begin{thebibliography}{10}
\providecommand{\url}[1]{#1}
\csname url@samestyle\endcsname
\providecommand{\newblock}{\relax}
\providecommand{\bibinfo}[2]{#2}
\providecommand{\BIBentrySTDinterwordspacing}{\spaceskip=0pt\relax}
\providecommand{\BIBentryALTinterwordstretchfactor}{4}
\providecommand{\BIBentryALTinterwordspacing}{\spaceskip=\fontdimen2\font plus
\BIBentryALTinterwordstretchfactor\fontdimen3\font minus \fontdimen4\font\relax}
\providecommand{\BIBforeignlanguage}[2]{{%
\expandafter\ifx\csname l@#1\endcsname\relax
\typeout{** WARNING: IEEEtran.bst: No hyphenation pattern has been}%
\typeout{** loaded for the language `#1'. Using the pattern for}%
\typeout{** the default language instead.}%
\else
\language=\csname l@#1\endcsname
\fi
#2}}
\providecommand{\BIBdecl}{\relax}
\BIBdecl

\bibitem{johnson2019survey}
J.~M. Johnson and T.~M. Khoshgoftaar, ``Survey on deep learning with class imbalance,'' \emph{Journal of big data}, vol.~6, no.~1, pp. 1--54, 2019.

\bibitem{shorten2021text}
C.~Shorten, T.~M. Khoshgoftaar, and B.~Furht, ``Text data augmentation for deep learning,'' \emph{Journal of big Data}, vol.~8, no.~1, p. 101, 2021.

\bibitem{sufi2024generative}
F.~Sufi, ``{Generative pre-trained transformer (GPT) in research: A systematic review on data augmentation},'' \emph{Information}, vol.~15, no.~2, p.~99, 2024.

\bibitem{chen2023empirical}
J.~Chen, D.~Tam, C.~Raffel, M.~Bansal, and D.~Yang, ``{An empirical survey of data augmentation for limited data learning in NLP},'' \emph{Transactions of the Association for Computational Linguistics}, vol.~11, pp. 191--211, 2023.

\bibitem{zakharova2021green}
O.~V. Zakharova, I.~N. Pupysheva, T.~Y. Payusova, A.~V. Zakharov, and L.~Sulkarnaeva, ``Green values in crowdfunding projects,'' \emph{Glocalism}, no.~1, 2021.

\bibitem{glazkova2022detecting}
A.~V. Glazkova, O.~V. Zakharova, A.~V. Zakharov, N.~N. Moskvina, T.~R. Enikeev, and A.~N. Hodyrev, ``{Detecting Mentions of Green Practices in Social Media Based on Text Classification},'' \emph{Modeling and Analysis of Information Systems}, vol.~29, no.~4, pp. 316--332, 2022.

\bibitem{zakharova2021greenA}
O.~V. Zakharova, T.~I. Payusova, I.~D. Akhmedova, and L.~G. Suvorova, ``Green practices: Ways to investigation,'' \emph{Sotsiologicheskie issledovaniya}, no.~4, pp. 25--36, 2021.

\bibitem{wei2019eda}
J.~Wei and K.~Zou, ``{EDA}: Easy data augmentation techniques for boosting performance on text classification tasks,'' in \emph{Proceedings of the 2019 Conference on Empirical Methods in Natural Language Processing and the 9th International Joint Conference on Natural Language Processing (EMNLP-IJCNLP)}, 2019, pp. 6382--6388.

\bibitem{karimi2021aeda}
A.~Karimi, L.~Rossi, and A.~Prati, ``{AEDA}: An easier data augmentation technique for text classification,'' in \emph{Findings of the Association for Computational Linguistics: EMNLP 2021}, 2021, pp. 2748--2754.

\bibitem{sennrich1511improving}
R.~Sennrich, B.~Haddow, and A.~Birch, ``Improving neural machine translation models with monolingual data,'' \emph{arXiv preprint arXiv:1511.06709}, 2015.

\bibitem{kumar2020data}
V.~Kumar, A.~Choudhary, and E.~Cho, ``Data augmentation using pre-trained transformer models,'' in \emph{Proceedings of the 2nd Workshop on Life-long Learning for Spoken Language Systems}, 2020, pp. 18--26.

\bibitem{wu2019conditional}
X.~Wu, S.~Lv, L.~Zang, J.~Han, and S.~Hu, ``Conditional {BERT} contextual augmentation,'' in \emph{Computational Science--ICCS 2019: 19th International Conference, Faro, Portugal, June 12--14, 2019, Proceedings, Part IV 19}.\hskip 1em plus 0.5em minus 0.4em\relax Springer, 2019, pp. 84--95.

\bibitem{dai2023auggpt}
H.~Dai, Z.~Liu, W.~Liao, X.~Huang, Y.~Cao, Z.~Wu, L.~Zhao, S.~Xu, W.~Liu, N.~Liu \emph{et~al.}, ``Aug{GPT}: Leveraging {ChatGPT} for text data augmentation,'' \emph{arXiv preprint arXiv:2302.13007}, 2023.

\bibitem{piedboeuf2023chatgpt}
F.~Piedboeuf and P.~Langlais, ``{Is ChatGPT the ultimate Data Augmentation Algorithm?}'' in \emph{Findings of the Association for Computational Linguistics: EMNLP 2023}, 2023, pp. 15\,606--15\,615.

\bibitem{zhao2024improving}
H.~Zhao, H.~Chen, T.~A. Ruggles, Y.~Feng, D.~Singh, and H.-J. Yoon, ``{Improving Text Classification with Large Language Model-Based Data Augmentation},'' \emph{Electronics}, vol.~13, no.~13, p. 2535, 2024.

\bibitem{yoo2021gpt3mix}
K.~M. Yoo, D.~Park, J.~Kang, S.-W. Lee, and W.~Park, ``{GPT3M}ix: Leveraging large-scale language models for text augmentation,'' in \emph{Findings of the Association for Computational Linguistics: EMNLP 2021}, 2021, pp. 2225--2239.

\bibitem{sarker2023medical}
S.~Sarker, L.~Qian, and X.~Dong, ``Medical data augmentation via {ChatGPT}: A case study on medication identification and medication event classification,'' \emph{arXiv preprint arXiv:2306.07297}, 2023.

\bibitem{cohen2023enhancing}
S.~Cohen, D.~Presil, O.~Katz, O.~Arbili, S.~Messica, and L.~Rokach, ``Enhancing social network hate detection using back translation and {GPT}-3 augmentations during training and test-time,'' \emph{information Fusion}, vol.~99, p. 101887, 2023.

\bibitem{shushkevich2023improving}
E.~Shushkevich, M.~Alexandrov, and J.~Cardiff, ``{Improving multiclass classification of fake news using BERT-based models and ChatGPT-augmented data},'' \emph{Inventions}, vol.~8, no.~5, p. 112, 2023.

\bibitem{wozniak2023big}
S.~Wo{\'z}niak and J.~Koco{\'n}, ``{From Big to Small Without Losing It All: Text Augmentation with ChatGPT for Efficient Sentiment Analysis},'' in \emph{2023 IEEE International Conference on Data Mining Workshops (ICDMW)}.\hskip 1em plus 0.5em minus 0.4em\relax IEEE, 2023, pp. 799--808.

\bibitem{moller2024parrot}
A.~G. M{\o}ller, A.~Pera, J.~Dalsgaard, and L.~Aiello, ``{The Parrot Dilemma: Human-Labeled vs. LLM-augmented Data in Classification Tasks},'' in \emph{Proceedings of the 18th Conference of the European Chapter of the Association for Computational Linguistics (Volume 2: Short Papers)}, 2024, pp. 179--192.

\bibitem{zakharova2024greenru}
O.~Zakharova and A.~Glazkova, ``{GreenRu: A {R}ussian Dataset for Detecting Mentions of Green Practices in Social Media Posts},'' \emph{Applied Sciences}, vol.~14, no.~11, p. 4466, 2024.

\bibitem{zakharova2022importance}
O.~V. Zakharova, A.~V. Glazkova, I.~N. Pupysheva, and N.~V. Kuznetsova, ``The importance of green practices to reduce consumption,'' \emph{Changing Societies \& Personalities. 2022. Vol. 6. Iss. 4}, pp. 884--905, 2022.

\bibitem{zmitrovich2024family}
D.~Zmitrovich, A.~Abramov, A.~Kalmykov, V.~Kadulin, M.~Tikhonova, E.~Taktasheva, D.~Astafurov, M.~Baushenko, A.~Snegirev, T.~Shavrina \emph{et~al.}, ``A family of pretrained transformer language models for {R}ussian,'' in \emph{Proceedings of the 2024 Joint International Conference on Computational Linguistics, Language Resources and Evaluation (LREC-COLING 2024)}, 2024, pp. 507--524.

\bibitem{clark2020electra}
\BIBentryALTinterwordspacing
K.~Clark, M.-T. Luong, Q.~V. Le, and C.~D. Manning, ``{ELECTRA}: Pre-training text encoders as discriminators rather than generators,'' in \emph{ICLR}, 2020. [Online]. Available: \url{https://openreview.net/pdf?id=r1xMH1BtvB}
\BIBentrySTDinterwordspacing

\bibitem{kuratov2019adaptation}
Y.~Kuratov and M.~Arkhipov, ``{Adaptation of deep bidirectional multilingual transformers for {R}ussian language},'' in \emph{Komp'juternaja Lingvistika i Intellektual'nye Tehnologii}, 2019, pp. 333--339.

\bibitem{devlin-etal-2019-bert}
J.~Devlin, M.-W. Chang, K.~Lee, and K.~Toutanova, ``{BERT}: Pre-training of deep bidirectional transformers for language understanding,'' in \emph{2019 Conference of the North {A}merican Chapter of the Association for Computational Linguistics: Human Language Technologies, Volume 1 (Long and Short Papers)}.\hskip 1em plus 0.5em minus 0.4em\relax Minneapolis, Minnesota: Association for Computational Linguistics, 2019, pp. 4171--4186.

\bibitem{rajapakse2019simpletransformers}
T.~C. Rajapakse, ``Simple transformers,'' \url{https://github.com/ThilinaRajapakse/simpletransformers}, 2019.

\bibitem{lin-2004-rouge}
\BIBentryALTinterwordspacing
C.-Y. Lin, ``{ROUGE}: A package for automatic evaluation of summaries,'' in \emph{Text Summarization Branches Out}.\hskip 1em plus 0.5em minus 0.4em\relax Barcelona, Spain: Association for Computational Linguistics, Jul. 2004, pp. 74--81. [Online]. Available: \url{https://aclanthology.org/W04-1013}
\BIBentrySTDinterwordspacing

\bibitem{zhangbertscore}
T.~Zhang, V.~Kishore, F.~Wu, K.~Q. Weinberger, and Y.~Artzi, ``{BERTS}core: Evaluating text generation with {BERT},'' in \emph{International Conference on Learning Representations}, 2020.

\end{thebibliography}

\end{document}